\documentclass[runningheads]{llncs} 

\usepackage{graphicx}
\usepackage{xcolor}
\usepackage{pbox}
\usepackage{array,makecell}
\usepackage{algorithmic} 
\usepackage{algorithm,algorithmic,amsmath}
\usepackage{subfigure}
\DeclareUnicodeCharacter{2212}{-}
\usepackage{multirow}
\usepackage{tablefootnote}

\setcellgapes{5pt}

\usepackage[labelsep=period]{caption}
\renewcommand{\thetable}{\textbf{\Roman{table}}}
\renewcommand\thesection{\Roman{section}}
\begin{document}
\title{IS-CAM: Integrated Score-CAM for axiomatic-based explanations}
%
%
\author{Rakshit Naidu\inst{1} \and
Ankita Ghosh\inst{1} \and
Yash Maurya\inst{1} \and
Shamanth R Nayak K\inst{1} \and
Soumya Snigdha Kundu\inst{2}}
\authorrunning{R. Naidu, A. Ghosh, Y. Maurya, S. R. Nayak K., S. S. Kundu}
%
\institute{Manipal Institute of Technology\\
\email{\{nemakallu.rakshit, ankita.ghosh1, yash.maurya1, shamanth.k\}@learner.manipal.edu}\\
\and
SRM Institute of Science and Technology\\
\email{sk7610@srmist.edu.in}}
\maketitle              
\begin{abstract}
Convolutional Neural Networks have been known as black-box models as humans cannot interpret their inner functionalities. With an attempt to make CNNs more interpretable and trustworthy, we propose IS-CAM (Integrated Score-CAM), where we introduce the integration operation within the Score-CAM pipeline to achieve visually sharper attribution maps quantitatively. Our method is evaluated on 2000 randomly selected images from the ILSVRC 2012 Validation dataset, which proves the versatility of IS-CAM to account for different models and methods.

\end{abstract}
\section{Introduction}
Convolutional Neural Networks (CNNs) are paramount when it comes to solving state of the art vision problems. 
The deployment of these models in sensitive situations such as the medical and security industry cannot be done without understanding and interpreting the intuition of the models as that will greatly increase the chances for model failure and deplete the confidence of the model.
To overcome these concerns and maintain the sensitivity of the task, 
a new research direction was put forward in order to build explainable models with CAMs \cite{zhou2015learning}.
Explainable models not only help in recognizing the drawbacks but also help in generating insights and accumulation of valuable information in tandem to the model's inference. It also helps in debugging the model and removing bias. 
Our work builds upon the CAM-based approaches \cite{Wang2020ScoreCAMSV} \cite{IntegratedGrad17} 
, which acquire attribution maps by a linear combination of the weights and the activation maps. While there are two different approaches to using CAMs we focused on the gradient-free approach as there are issues pertaining to gradient CAMs such as saturation and false confidence \cite{selvaraju2016gradcam}. 
One of the first approaches towards a gradient-free method was Score-CAM \cite{Wang2020ScoreCAMSV}
, but due to its coarse localization, it tends to lead to erratic localizations in certain cases. Our contributions to overcome the existing issues are:

\begin{itemize}
    \item We propose a new axiomatic-based approach IS-CAM, which is combined within the Score-CAM pipeline to produce sharper attribution maps.
    \item We attain improved performance in comparison to previous CAM-based methods. We quantitatively evaluate over faithfulness and localization tasks, which indicate better localized decision-related features of IS-CAM.
\end{itemize}

\section{Related Work}

\textbf{IntegratedGrad: } \cite{IntegratedGrad17} demonstrated their ability to debug a network by extracting certain rules from the network, thereby enabling the users to engage more with the models and understand the network's predictions. They introduced two axioms for attribution methods, namely: \textit{Sensitivity} (if there is a feature difference between the input and the baseline and have different predictions, then the differing feature should be assigned a non-zero attribution) and \textit{Implementation Invariance} (if two networks give the same output for all inputs, despite having different implementations, the attributions should be equal in these two \textit{functionally equivalent} networks). 
The Integrated gradient along the ${i}^{th}$ dimension is denoted by:
\begin{equation}
    \left( x_{i}-x_{i}'\right) \times \int ^{1}_{\alpha =0}\dfrac {\partial F( x' +\alpha \times \left( x-x'\right)) }{\partial x_{i}}d\alpha
\end{equation}

where ${x}$ is the input and ${x'}$ is the baseline. $\dfrac {\partial F\left( x\right) }{\partial x_{i}}$ represents the gradient of ${F(x)}$ along the ${i}^{th}$ dimension.

\textbf{Class Activation Maps: } 
The inspiration driving CAM \cite{zhou2015learning} is that each activation map $A^k_l$, where $A$ denotes the activation map for the $k$-th channel and $l$-th layer, contains distinctive spatial information about the input $X$. 
For a given class $c$, the input to the softmax $S_c$ is $\sum\limits_{k}w_c^kA_l^k$ where $w_c^k$ is the weight corresponding to class $c$ for $k$-th layer and $A_l^k$ represents the global pooling layer.
CAM $L^c_{CAM}$ can be defined as
\begin{equation}
    L^c_{CAM}= ReLU\left(\sum\limits_{k}w_c^kA_{l-1}^k\right)
\end{equation}
\newline

\textbf{Grad-CAM: }
As CAM is limited to GAP-based CNN models, Grad-CAM \cite{selvaraju2016gradcam} was developed to generalize for a wider range of CNN architectures. To obtain each neuron for a decision of interest, Grad-CAM uses the gradient information flowing into the last convolutional layer. Considering an activation map $A^{k}$ for the $k$-th channel, Grad-CAM $L^{c}_{Grad-CAM}$ for target class $c$ can be defined as




\begin{equation}
    L^c_{Grad−CAM}=ReLU\left(\sum\limits_{k} \alpha_{c}^{k} A^{k}\right)
\end{equation}
where $\alpha_c^k$ represents the neuron importance weights. $\alpha_c^k=\frac{1}{Z}\sum\limits_{i}\sum\limits_{j}\frac{\partial Y_c}{\partial A^k_{i j}}$ where $Y_c$ is the score computed for the target class, $(i,j)$ represents the location of the pixel and $Z$ denotes the total number of pixels.

Some other variations of Grad-CAM like Grad-CAM++ and Smooth Grad-CAM++ serve as a comparison for our algorithm in the sections that follow.
\newline

\textbf{Score-CAM: }
In Score-Cam \cite{Wang2020ScoreCAMSV}, the weights of the score obtained for a specific target class $c$ are utilized. Score-CAM disposes of the reliance on the gradient and provides a more generalized framework as it only requires access to the class activation map and output scores. Considering an activation map $A^k_l$ for $k$-th channel and $l$-th convolutional layer, Score-CAM $L^{c}_{Score-CAM}$ can be defined as
\begin{equation}
    L^c_{Score−CAM}=ReLU\left(\sum\limits_{k} \alpha_{c}^{k} A^{k}_{l}\right)
\end{equation}
where $\alpha_{c}^{k}$ denotes the channel-wise Increase of Confidence performed on  $A^{k}_{l}$ in order to measure the importance of the activation map.

\section{Proposed Approach}

\begin{figure*}[t]
\centering
    \includegraphics[width=12cm]{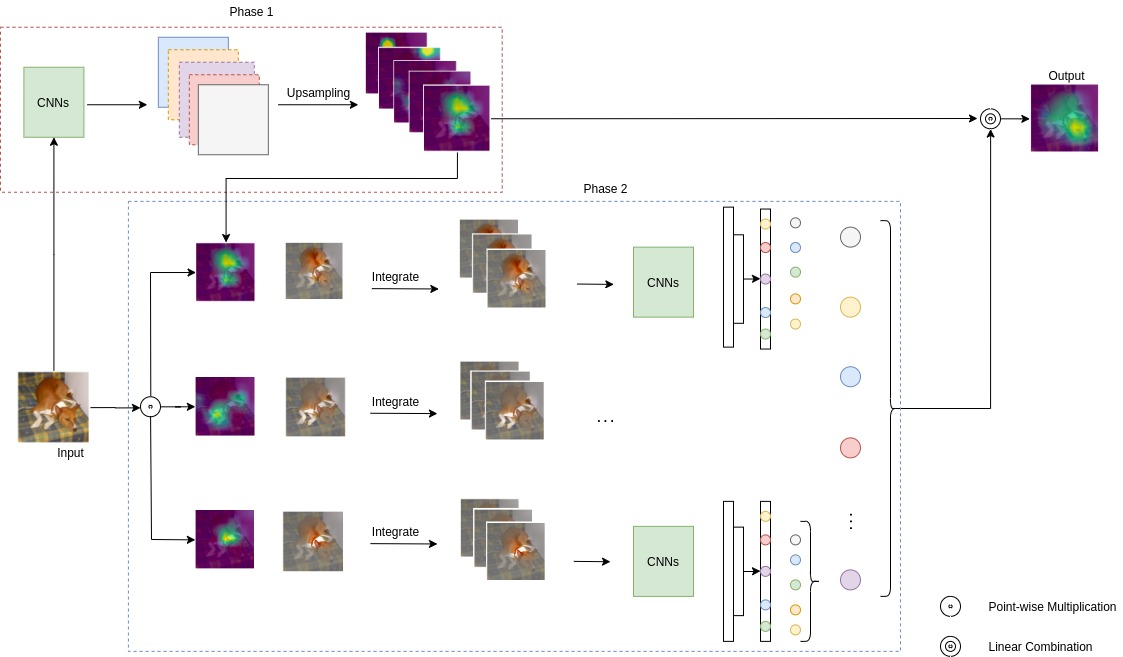}
    \caption{Pipeline of the proposed IS-CAM approach. The saliency map is produced by the linear combination of the average scores after "integration" and the upsampled activation maps. The average score is obtained from performing summation over the normalized input mask at every interval. }
    \label{fig:IS-CAM_pipeline}
\end{figure*}

In this section, we explain our approach on how we combine IntegratedGrad \cite{IntegratedGrad17} within the Score-CAM pipeline. Figure~\ref{fig:IS-CAM_pipeline} shows our pipeline.

We set a parameter $N$ as the number of intervals between the range [0, 1]. As the integration operation is analogous to the summation operation, we calculate scores of the maps at each step of the interval from $0$ to $1$. Finally, we calculate the average of the scores generated as the mean operation is sensitive to changes in the saliency maps generated at each step of the process. Note that $M_{0} = 0$.

\textbf{Integrating over the input mask}:

\begin{equation}
L^{c}_{IS-CAM} = ReLU\left( \sum _{k}\alpha ^{c}_{k}A^{k}_{l}\right)
\end{equation}
$where$
\begin{equation}
    \alpha^{c}_{k} = \dfrac{\sum ^{N}_{i=1}\left(C(M_{i})\right)}{N}
\end{equation}
\begin{equation}
    M_{i+1} \leftarrow M_{i} + \left((X_{0} * A^{k}_{l}) * \frac{i}{N}\right)
\end{equation}

\textbf{Normalization}: 

As the spatial region needs to focused on the object in the image, we leverage the features within a particular region by following the same normalization function as stated in \cite{Wang2020ScoreCAMSV}, \cite{wang2020sscam}. The normalization used in the algorithm is given as:

\begin{equation}
s\left(A^{k}_{l}\right) = \dfrac{A^{k}_{l} - min(A^{k}_{l})}{max(A^{k}_{l}) - min(A^{k}_{l})}
\end{equation}

\section{Experiments}
\renewcommand\thesubsection{\thesection.\alph{subsection}}

In this section, we conduct experiments to evaluate the effectiveness of the proposed explanation method. Our setup is similar to that described in \cite{chattopadhyay2017gradcam}, \cite{petsiuk2018rise}, \cite{Wang2020ScoreCAMSV}. First, a qualitative output comparison of the architectures by visualization on the  ILSVRC 2012 Validation set in section A. Second, we assess the fairness of the interpretations of architectures for object recognition in section B. Third, the Energy-based pointing game (proposed in \cite{chattopadhyay2017gradcam}) is used to evaluate the bounding boxes for the class-conditional object localization in a given image in section C over 2000 uniformly random selected images from the ILSVRC Validation Set 2012. 

Our comparative analysis extends to five other known CAM methods, Grad-CAM \cite{selvaraju2016gradcam}, Grad-CAM++ \cite{chattopadhyay2017gradcam}, Smooth Grad-CAM++ \cite{smoothgradcam++} 
Score-CAM \cite{Wang2020ScoreCAMSV}, and Smoothed Score-CAM \cite{wang2020sscam}. The images are resized with a fixed size (224, 224, 3), condensed into the [0,1] range and then, normalized using ImageNet \cite{ImageNet} weights (mean vector : [0.485, 0.456, 0.406] and standard deviation vector [0.229, 0.224, 0.225]). For simplicity, baseline image $X_{b}$ is set to 0(as shown in Channel-wise Increase in Confidence \cite{Wang2020ScoreCAMSV}).
\newline

A. \emph{Visual Comparison}

\begin{figure*}[ht]
    \includegraphics[width=12cm]{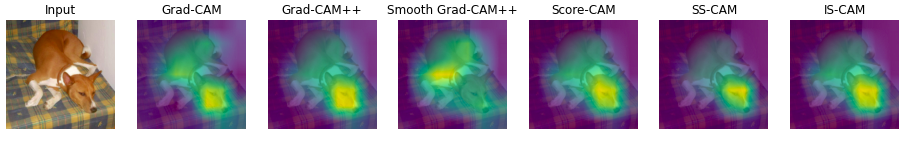}
    \includegraphics[width=12cm]{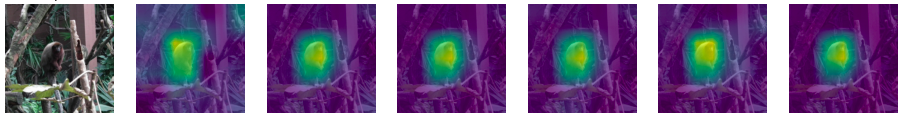}
    \includegraphics[width=12cm]{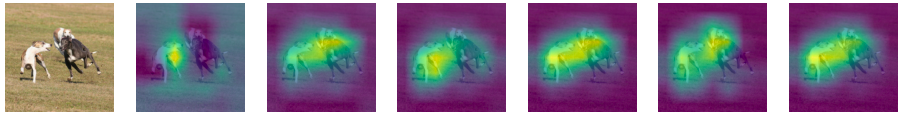}
    \caption{Depicts the Imagenet Labels (Row-wise): Basenji, Capuchin and Whippet. 
    This figure is used for a Visual Comparison of our approach with the other existing approaches. We use $N=10$ here.}
    \label{fig:IS-CAM_visual_comparison}
\end{figure*}

\indent To perform this experiment, 2,000 images were randomly selected from the 2012 ILSVRC Validation Set. Fig \ref{fig:IS-CAM_visual_comparison} shows a few photos comparing our approach to prevailing CAM approaches. Here, we used N = 15 and $\sigma$ = 2 for SS-CAM. Even though we achieve comparable visual results to Score-CAM, we perform better quantitatively in terms of the Faithfulness explanations as shown in the next section.

B. \emph{Faithfulness Evaluations}

\indent Faithfulness evaluations are carried out as described in Grad-CAM++ \cite{chattopadhyay2017gradcam} for the purpose of Object Recognition. Three metrics called Average Drop, Average Increase In Confidence, and Win $\%$ are implemented.
These metrics are tested for 2000 images randomly chosen from the ILSVRC 2012 Validation set, using the pre-trained VGG-16 model. To perform this sub-experiment, we used N = 15 and $\sigma$ = 2 (for SS-CAM).

\begin{table}
\caption{Average AUC scores of the Insertion curve(the higher, the better) and Deletion curve(the lower, the better) over all the 2000 images.}
\label{tab2}
\vspace{0.3cm}
\centering

\begin{tabular}{m{3cm}
>{\centering\arraybackslash}m{3cm}
>{\centering\arraybackslash} m{3cm} >{\centering\arraybackslash} m{3cm} 
}
\hline

\textbf{CAM techniques} & \thead{\textbf{Insertion \%}} & \thead{\textbf{Deletion \%}}\\

\hline
\textbf{Grad-CAM} & 45.25 & 11.25\\
\hline
\textbf{G-CAM ++ \tablefootnote{Grad-CAM++}} & 44.94 & 11.41\\
\hline
\textbf{SG-CAM++ \tablefootnote{Smooth Grad-CAM++}} & 42.68 & 13.43\\
\hline
\textbf{Score-CAM} & \textbf{48.22} & \textbf{9.92}\\
\hline
\textbf{SS-CAM} & 45.92 & 11.46\\
\hline
\textbf{IS-CAM} & 48.13 & \textbf{9.92}\\

\hline
\end{tabular}
\end{table} 

\begin{figure*}[t]
\centering
    \includegraphics[width=8cm]{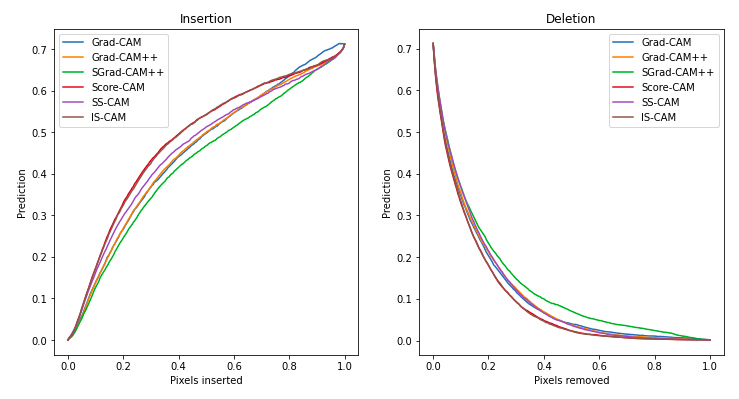}
    \caption{Insertion and Deletion curve charts for Table~\ref{tab2}.}
    \label{fig:-Insertion_deletion}
\end{figure*}

\indent
Insertion and Deletion Curves are used to calculate the Area Under Curve (AUC) metric to understand how many pixels of the saliency map will either add or reduce the scores of the resulting fractioned maps. We average the resulting pixel values at each stage(deleting/inserting 224 pixels) over all the 2000 images and produce graphs in Figure~\ref{fig:-Insertion_deletion}. The Deletion operation demonstrates the ability to remove the map information pixel-wise. A sharp decline and a lower AUC of the generated scores imply a good explanation.  The Insertion operation evaluates the ability to reconstruct the saliency map from a given baseline. A sharp rise and higher AUC of the generated scores imply a good explanation. 

\begin{enumerate}
    \item \textit{Average Drop \%}: The Average Drop refers to the maximum positive difference in the predictions made by the prediction using the input image and the prediction using the saliency map. It is given as: $\sum_{i=1}^{N} \frac{max(0, Y_{i}^{c} - O_{i}^{c})}{Y_{i}^{c}} \times 100$. Here, \textit{Y\textsubscript{i}\textsuperscript{c}} refers to the prediction score on class \textit{c} using the input image \textit{i} and \textit{O\textsubscript{i}\textsuperscript{c}} refers to the prediction score on class \textit{c} using the saliency map produced over the input image \textit{i}.
    
    \item \textit{Increase in Confidence \%}: The Average Increase in Confidence is denoted as: $\sum_{i=1}^{N} \frac{Fun(Y_{i}^{c} < O_{i}^{c})}{N} \times 100$
     where \textit{Fun} refers to a boolean function which returns 1 if the condition inside the brackets is true, else the function returns 0. The symbols are referred to as shown in the above experiment for Average Drop.
     
     \item \textit{Win \%}: The Win percentage refers to the decrease in the model's confidence for an explanation map generated by IS-CAM to the confidence generated by another algorithm. This metric is compared to the confidence generated by SS-CAM \cite{wang2020sscam} maps and Score-CAM \cite{Wang2020ScoreCAMSV} maps with IS-CAM maps. When our approach is compared to SS-CAM, we get 59.25\% and when compared to Score-CAM, we get 52.35\% using VGG-16(higher is better); which indicates that IS-CAM performs better with respect to this metric.
    
\end{enumerate}

The AUC scores, Average Drop and Increase in Confidence indicate that IS-CAM performs better on an overall perspective. While Score-CAM performs well in AUC scores it fails to do so in Average Drop and Inc\% using VGG-16 . Likewise, SS-CAM does well in Average Drop and Inc\% but it fails to do so in AUC scores. IS-CAM does well in both perspectives which shows its profound versatility. 

\begin{table}[h]
 \centering
\caption{Average Drop (the lower, the better) and Average Increase in Confidence (the higher, the better) across 2000 ILSVRC Validation images.}
\vspace{0.3cm}
    
    \begin{tabular}{m{2.2cm} | c c | c c | c c}
    
\hline

\textbf{CAM} & \multicolumn{2}{c}{\textbf{VGG-16}} & \multicolumn{2}{c}{\textbf{Resnet}} & \multicolumn{2}{c}{\textbf{SqueezeNet}} \\
\cline{2-7}
\textbf{Techniques} & {Avg Drop\%} & Avg Inc\% & Avg Drop\% & Avg Inc\% & Avg Drop\% & Avg Inc\% \\
\hline

\textbf{Score-CAM} & 66.03 & 51.85 & 64.23 & 53.55 & 13.42 & 60.85\\
\hline
\textbf{SS-CAM} & 79.15 & 51.30 & 64.53 & \textbf{54.80} & \textbf{12.06} & \textbf{64.85}\\
\hline
\textbf{IS-CAM} & \textbf{63.30} & \textbf{52.35} & \textbf{64.85} & 53.50 & 13.00 & 62.15\\
\hline

    \end{tabular}
    
    \label{tab:my_label}
   
\end{table}







C. \emph{Localization Evaluations}

\indent This section accomplishes evaluations related to Bounding boxes. A metric known as Energy-based pointing game, as introduced in \cite{Wang2020ScoreCAMSV}, is employed for our localization experiments. This helps in calculating how much energy of the saliency map falls within the given Bounding box. This is achieved in two steps. The first step of  this is where the input image is binarized, specifically with the interior of the Bounding box marked as 1 and the region outside the Bounding box as 0. This is then multiplied element-wise with the saliency map generated for the input image and summed over to calculate proportion ratio which is given as - $Proportion = \frac{\sum L^{c}_{(i,j) \in bbox} }{\sum L^{c}_{(i,j)\in bbox}+\sum L^{c}_{(i,j)\notin bbox}}$. We evaluate this metric on 2000 randomly selected images from the ILSVRC 2012 Validation set \cite{ImageNet}. These images are then fed to 3 pre-trained models, namely,
VGG-16 \cite{Simonyan2015VeryDC}, ResNet-18(Residual Network with 18 layers) \cite{He2016DeepRL}, and SqueezeNet1.0 \cite{Iandola2017SqueezeNetAA}. Table~\ref{tab3} portrays the results of the localization evaluation for the 3 architectures.
We see that IS-CAM performs better than most techniques in all three models. It also achieves the highest value for the VGG-16 variant.

\begin{table}
\caption{Localization Evaluation}
\label{tab3}

\centering
\vspace{0.3cm}
\begin{tabular}{m{3cm}
>{\centering\arraybackslash}m{3cm}
>{\centering\arraybackslash} m{3cm} >{\centering\arraybackslash} m{3cm} 
}
\hline

\textbf{CAM techniques} & \thead{\textbf{VGG-16} \\\textbf{Proportion(\%)}} & \thead{\textbf{ResNet18} \\ \textbf{Proportion(\%)}} & \thead{\textbf{SqueezeNet1.0} \\\textbf{Proportion(\%)}}\\

\hline
\textbf{Grad-CAM} & 42.69 & 43.55 & 42.01\\
\hline
\textbf{G-CAM++} & 42.87 & 43.53 & 41.83\\
\hline
\textbf{SG-CAM++} & 42.97 & \textbf{43.56} & 41.77\\
\hline
\textbf{Score-CAM} & 43.07 & 43.46 & \textbf{42.48}\\
\hline
\textbf{SS-CAM} & 42.46 & 43.30 & 41.98\\
\hline
\textbf{IS-CAM} & \textbf{43.17} & 43.52 & 42.40\\

\hline
\end{tabular}
\end{table}









\section{Conclusion \& Future Work}
Our proposed method involves integrating over the input mask and averaging the scores obtained from the normalised masks.
According to our experiments, the increase or decrease of the value $N$, does not have a significant impact on the visual attribution map produced. The effect of $N$ is quite evident quantitatively as demonstrated in our experiments. In the future,
we hope to test our algorithms in the medical domain to prove its effectiveness in sensitive real world scenarios.

\section*{Acknowledgment}
We thank Mr. Haofan Wang from Carnegie Mellon University for his valuable inputs during the discussion.
We would also like to thank the Research Society MIT, Manipal(RSM) for supporting and moderating the project.


%
%
%
\bibliographystyle{splncs04}
\bibliography{iscam}

\end{document}